\documentclass[runningheads]{llncs}
\usepackage[T1]{fontenc}
\usepackage{graphicx}
\usepackage{multirow}
\usepackage{amsmath}
\usepackage{subcaption}
\usepackage{booktabs}
\usepackage{makecell}
\usepackage{dirtytalk}
\begin{document}
\title{Precision Where It Matters: A Novel Spike Aware Mixed-Precision Quantization Strategy for LLaMA-based Language Models}
\titlerunning{Precision Where It Matters}
%
\author{Lucas Maisonnave\inst{1} \and
Cyril Moineau\inst{1} \and
Olivier Bichler\inst{1} \and
Fabrice Rastello\inst{2}}
\authorrunning{L.Maisonnave et al.}
%
\institute{Univ. Paris-Saclay, CEA LIST, F-91120, Palaiseau, France  \\
\email{{lucas.maisonnave, cyril.moineau, olivier.bichler}@cea.fr} \and
Univ. Grenoble Alpes, Inria, CNRS, Grenoble INP, LIG, 38000 Grenoble, France \\
\email{fabrice.rastello@inria.fr}\\}
\maketitle              
\begin{abstract}
Large Language Models (LLMs) have demonstrated remarkable capabilities in various natural language processing tasks. However, their size presents significant challenges for deployment and inference. This paper investigates the quantization of LLMs, focusing on the LLaMA architecture and its derivatives. We challenge existing assumptions about activation outliers in LLMs and propose a novel mixed-precision quantization approach tailored for LLaMA-like models.
Our method leverages the observation that activation spikes in LLaMA architectures are predominantly concentrated in specific projection layers. By applying higher precision (FP16 or FP8) to these layers while quantizing the rest of the model to lower bit-widths, we achieve superior performance compared to existing quantization techniques.
Experimental results on LLaMA2, LLaMA3, and Mistral models demonstrate significant improvements in perplexity and zero-shot accuracy, particularly for 8-bit per-tensor quantization. Our approach outperforms general-purpose methods designed to handle outliers across all architecture types, highlighting the benefits of architecture-specific quantization strategies.
This research contributes to the ongoing efforts to make LLMs more efficient and deployable, potentially enabling their use in resource-constrained environments. Our findings emphasize the importance of considering model-specific characteristics in developing effective quantization pipelines for state-of-the-art language models by identifying and targeting a small number of projections that concentrate activation spikes.

\end{abstract}
\section{Introduction}

Large Language Models (LLMs), such as GPT-4 and LLaMA, have become central to numerous natural language processing (NLP) tasks, demonstrating exceptional capabilities in understanding and generating human language. These models are typically built upon billions of parameters (and even trillions for GLaM \cite{du2022glam}), enabling them to capture complex linguistic patterns and nuances. However, the sheer size of LLMs brings significant challenges in terms of computational resources and storage requirements. The most recent models like GPT4-o or Claude 3.5 Sonnet are build with the objective to have a faster and lighter model to use on a phone in the future.

Compression methods like quantization is a very effective way to reduce model size, but also to accelerate inference. Various method were built for LLMs and work very efficiently like GPTQ \cite{frantar2022gptq} which is a Post-Training Quantization (PTQ) method able to reduce the size of weights up to 4 bits without significant drop of performance. Quantization Aware Training (QAT) is more difficult to apply because of the need to retrain the model from scratch but it can reduce memory footprint even further like BitNet 1.58 bits \cite{ma2024era} which is able to use ternary weights for a LLaMA architecture.

Today the urge to compress LLM models is motivated to reduce the models storage and increase their speed but also to reduce their environmental impact. Most of the newest open source models like LLaMA \cite{touvron2023llama}, BLOOM \cite{le2023bloom} or Mistral \cite{jiang2023mistral} are published with their carbon cost of training. For example the latest LLaMA3-70B model published by Meta consumed 1900 teqCO2 to be trained \cite{Meta_Llama_3_8B}. Compression and quantization are effective ways to reduce the energy consummation of LLM like QLoRA \cite{dettmers2024qlora} which enables to fine-tune an 8 bits quantized model with only one GPU A100.

While quantizing weights is relatively straightforward, quantizing activations presents a unique challenge, especially in LLMs. Activations often exhibit outliers—extremely high compared to the typical range \cite{yang2024mitigating,sun2024massive}. These outliers can distort the scale used for quantization, leading to significant information loss and degradation in model performance. The presence of outliers necessitates more sophisticated quantization strategies to ensure that the quantized activations retain the representational power of the original model.

Our research offers two primary contributions to LLM activation quantization:

\begin{itemize}
    \item We critically examine existing literature, challenging some overly broad assumptions about outlier behavior in LLMs. Specifically, we question the notion that certain dimensions consistently capture most outliers. We also highlight contradictions in reported outlier locations and imprecise outlier definitions, which hinder result reproducibility. This analysis leads us to conclude that developing a universal method for managing outliers across diverse decoder architectures like LLaMA, OPT, or GPT is challenging.
    \item Given these findings, we narrow our focus to the LLaMA architecture, an open-source, high-performance model that forms the foundation for various other architectures, including Mistral. Our experiments reveal the benefits of applying varied precision levels to specific model layers. We find that enhancing computational accuracy for certain layers yields significant advantages in the quantization process.
\end{itemize}

This approach allows us to address the unique characteristics of LLaMA-based models while developing an effective quantization strategy.

\section{Related Works}

\subsection{Quantization}
Quantizing model is the process of reducing the number of bits used to store and compute the model activations. To do this we need to define a scaling factor which defines the distance between 2 bins and the range of values to compress. For a symmetric uniform quantization we apply a rounding function on a scaled distribution :

\begin{equation}
\label{eq:quant}
    \hat{X} = \left \lfloor \frac{X}{\Delta} \right \rceil \Delta, \quad \Delta = \frac{\text{max}|X|}{2^{b-1} - 1}
\end{equation}

Where $\Delta$ is the scaling factor, $b$ is the bitwidth and we use the maximum absolute value as the range of our distribution which preserves extreme values for activations. Such quantization can be applied per-token so each token has a different scaling factor but it is more difficult to use efficiently in practice compared to per-tensor quantization which uses only one scaling factor for each activation tensor.
Scaling factors can be static at inference and based on statistics computed on a small part of the dataset or it can be dynamic and recomputed every time. Quantization can cause a huge drop of performance when it is done post-training \cite{xiao2023smoothquant,son2024prefixing} that's why some methods adapt weights to the noise added during a training phase \cite{wang2023bitnet,ma2024era}.

Usually for LLM we only quantize linear layers gathering most of the computation cost and we do not quantize the normalisation layer, the matrix multiplication and the softmax function inside the attention block.

\subsection{Outliers in LLMs}

Quantizing LLMs weights is quite easy and it doesn't require much efforts to make it work. Numerous techniques, such as GPTQ \cite{frantar2022gptq}, allow for 8-bit quantization without requiring retraining or sacrificing model accuracy. Some QAT methods can go even further and perform 1 bits quantization like BitNet \cite{wang2023bitnet} or ternary quantization \cite{ma2024era}. 

LLMs are challenging to quantize due to the widespread presence of extreme high values throughout the model's activations \cite{he2024understanding}. The scaling factor, which is directly related to the maximum absolute value, causes most of the distribution to be rounded to zero, leading to performance degradation. To address this issue, techniques like LLM.int8() \cite{dettmers2022gpt3} cluster these outliers and quantize them separately from the main distribution. Alternative methods such as SmoothQuant \cite{xiao2023smoothquant} shift the quantization challenge from activations to weights by introducing a scaling parameter between them. Other approaches attempt to relocate these spikes into sink tokens before quantization \cite{son2024prefixing}.

Ultimately, some research aims to study the problem upstream during the learning process and understand the causes of these \say{spikes} appearance, thus allowing to limit their impact after training \cite{yang2024mitigating}. Other works seek to better locate these outliers, particularly by visualizing the layers, dimensions, and tokens that could be their cause \cite{paglieri2024outliers,he2024understanding}.

\begin{figure}[ht]
    \begin{subfigure}[b]{0.49\textwidth}
         \centering
         \includegraphics[width=0.9\textwidth]{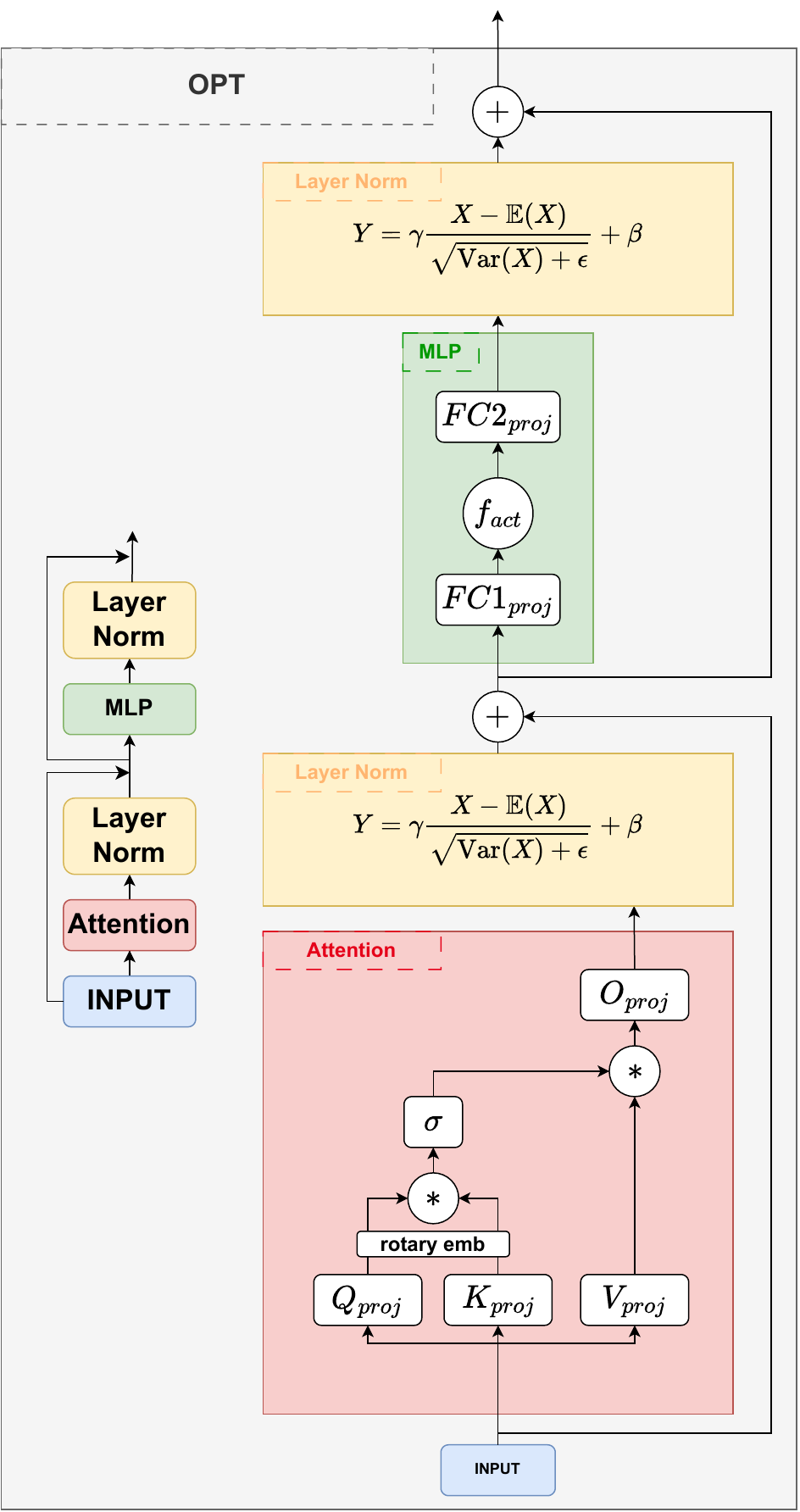}
         \caption{OPT}
         \label{fig:OPT}
     \end{subfigure}
     \hfill
     \begin{subfigure}[b]{0.49\textwidth}
         \centering
         \includegraphics[width=0.9\textwidth]{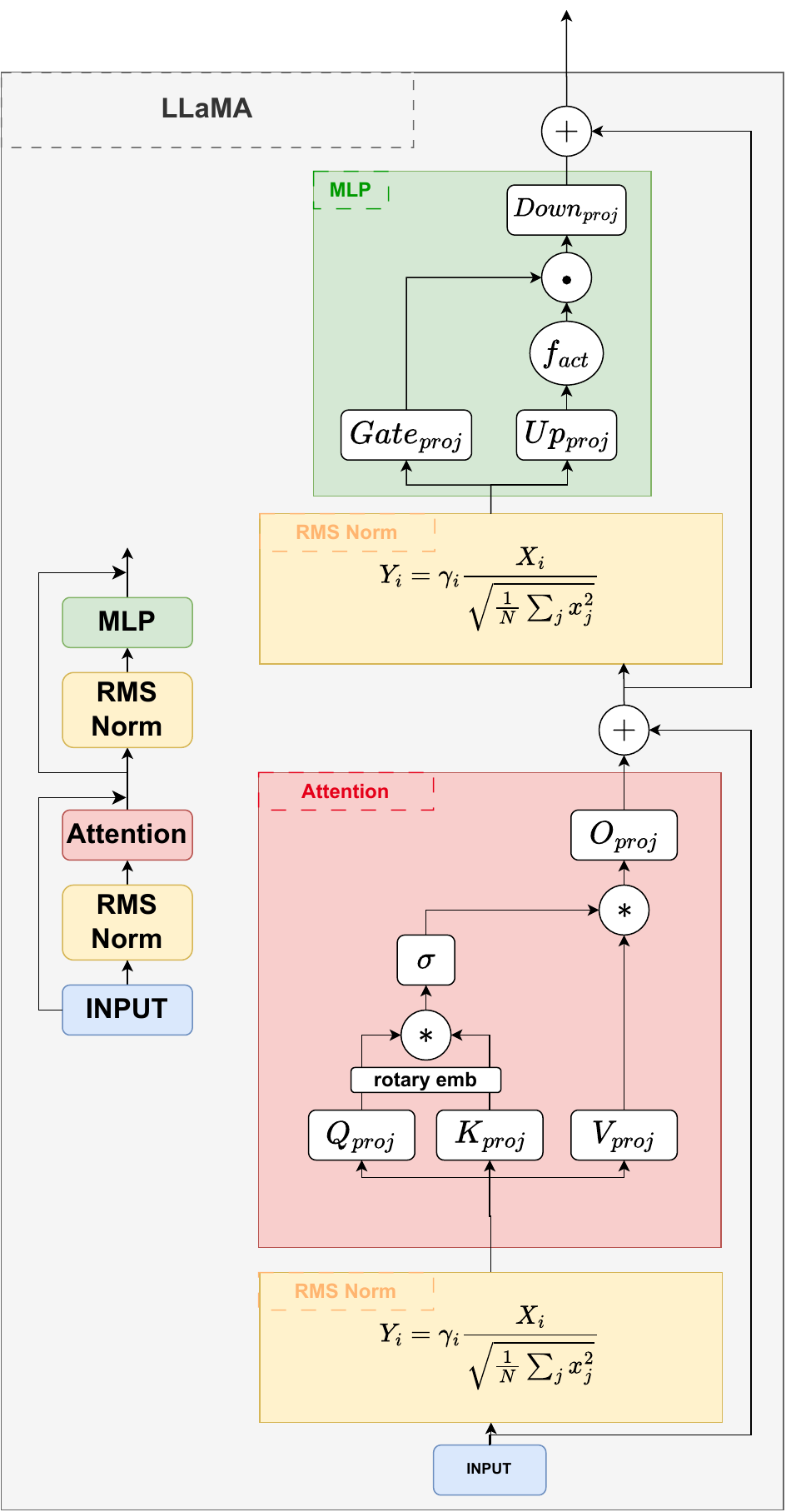}
         \caption{LLaMA}
         \label{fig:LLaMA}
     \end{subfigure}
     \caption{Causal architecture of OPT and LLaMA model. There are 2 main differences : the location of the layer norm and RMS Norm, and the MLP in LLaMA uses a gate projection }
        \label{fig:Architectures}
\end{figure}

\section{Analyzing Spikes in LLaMA Architectures}

This section aims to review and analyze the findings on activation outliers presented in current research, with the goal of extracting valuable insights to develop our own optimized pipeline. While numerous papers address activation outliers in LLMs, we've observed that some hypothesis about these outliers can be highly specific to the architecture studied. Notably, we found that outliers manifest quite differently in LLaMA-like architectures compared to OPT-like models (Figure \ref{fig:Architectures}), and appear to be more manageable in the former. Consequently, we've chosen to focus on the LLaMA architecture, developing a PTQ mixed precision method using FP8 format.

\subsection{Definition of an outlier}

The concept of outliers in transformer activations was first introduced in \cite{bondarenko2021understanding}. They defined outliers as values exceeding six standard deviations from the activation tensor's mean, using this definition to identify dimensions with high values in a BERT model.

LLM.int8() \cite{dettmers2022gpt3} took a different approach, eschewing a statistical definition in favor of problem-specific criteria: "features with a magnitude of at least 6.0, affecting a minimum of 25\% of layers and 6\% of sequence dimensions".

These definitions have evolved over time. Currently, an outlier is generally considered a value that is an order of magnitude larger than the main distribution. While less precise, this definition is more practical for our goal of reducing a distribution's maximum absolute value to achieve a more accurate scaling factor (\ref{eq:quant}).

In the context of quantization, what we refer to as an "outlier" is more accurately described as a "spike", a term introduced in \cite{yang2024mitigating,sun2024massive}. We will use 'spike' for the remainder of this paper when discussing outliers.

\begin{figure}[ht]
    \begin{subfigure}[b]{0.49\textwidth}
         \centering
         \includegraphics[width=\textwidth]{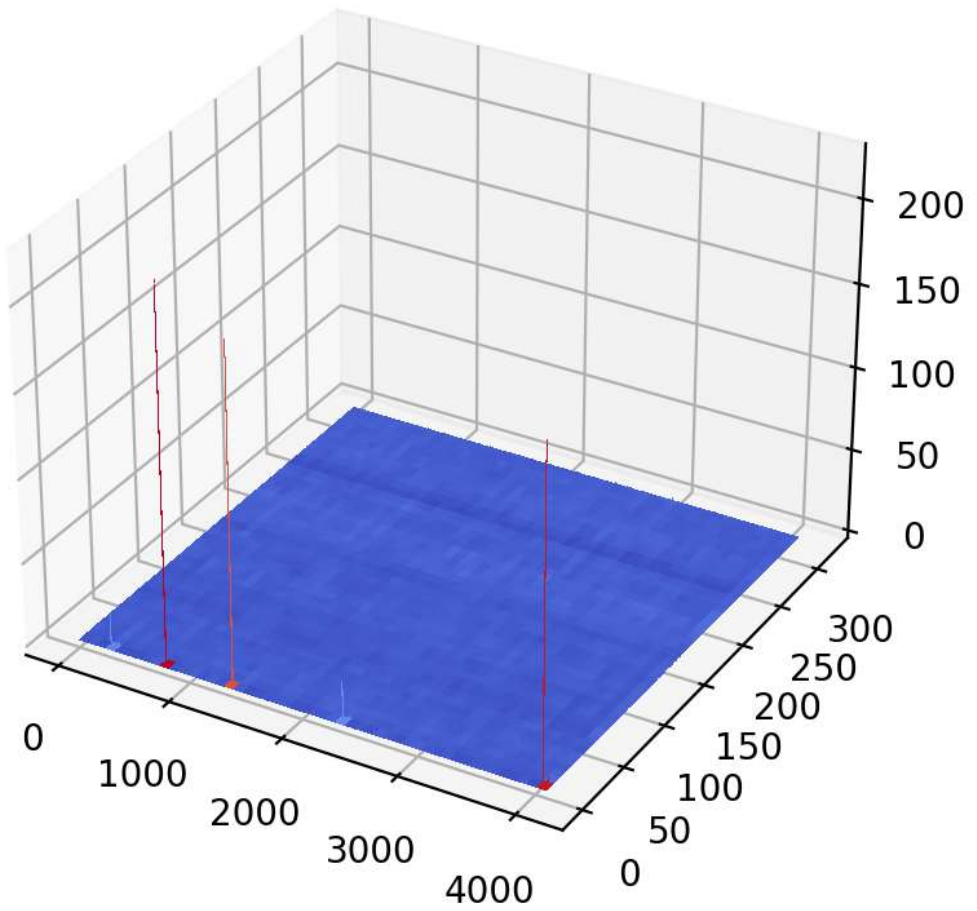}
         \caption{LLaMA3 8B}
         \label{fig:LLaMA8B_input_layernorm}
     \end{subfigure}
     \hfill
     \begin{subfigure}[b]{0.49\textwidth}
         \centering
         \includegraphics[width=\textwidth]{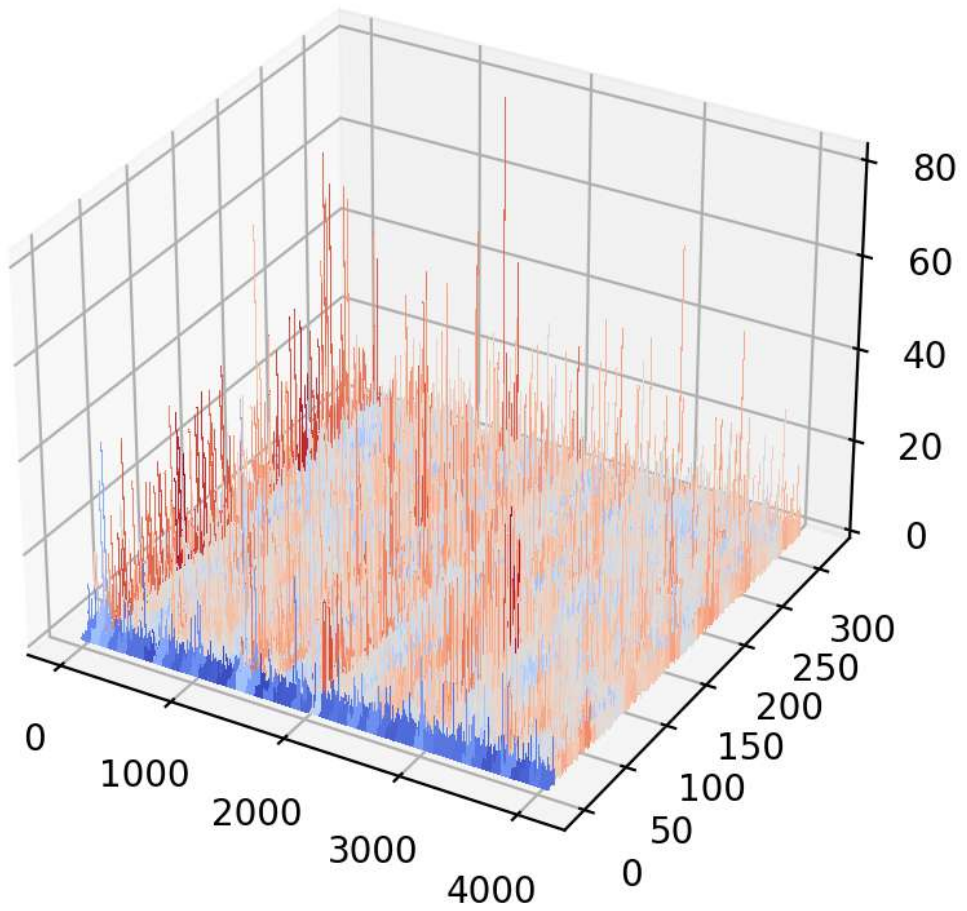}
         \caption{Mistral 7B}
         \label{fig:Mistral7B_input_layernorm}
     \end{subfigure}
     \caption{Layer 19 activations values of (a) the first RMSNorm's input in a LLaMA3-8B (b) k-proj output in a Mistral 7B}
        \label{fig:Activations_LN}
\end{figure}

\subsection{Dimensions gathering outliers}

A significant hypothesis in the literature posits that certain dimensions of the embedding absorb most of the model's outliers \cite{bondarenko2021understanding,dettmers2022gpt3,xiao2023smoothquant,yang2024mitigating}. Various approaches have been used to demonstrate this hypothesis, employing the different outlier definitions mentioned earlier. Recent studies have attempted to verify this claim on newer models beyond BERT or OPT \cite{paglieri2024outliers,he2024understanding}.

However, we question the universality of this statement, particularly when some findings are challenging to reproduce, such as in \cite{ahmadian2023intriguing} where a clear outlier definition is lacking. Our observations of LLaMA2 and LLaMA3 models reveal no dimensions consistently gathering outliers. Instead, we note spikes at specific points in activations (Figure \ref{fig:LLaMA8B_input_layernorm}). This contrasts with \cite{xiao2023smoothquant}, where one channel exhibits spikes across all tokens.

Interestingly, in the Mistral7B model, we do observe spikes across dimensions (Figure \ref{fig:Mistral7B_input_layernorm}), but these differ significantly from those reported in SmoothQuant or LLM.int8(). These patterns are consistent across all layers of LLaMA3 and Mistral7B, though the manifestation of outliers varies depending on the specific block examined (e.g., projection, layer normalization).

Our research corroborated a significant phenomenon in the LLaMA architecture: the presence of notable spikes in the Beginning-of-Text (BOT) token, as illustrated in Figure \ref{fig:LLaMA8B_input_layernorm}. To validate this observation, we conducted experiments where we selectively avoided quantizing this token and compared the resulting perplexity across various models (see Table \ref{tab:BOT_exp}). The results for LLaMA models showed a substantial reduction in perplexity when the BOT token was left unquantized, strongly indicating that these spikes are predominantly associated with this token. In contrast, while Mistral architectures also exhibited a decrease in perplexity under similar conditions, the effect was notably less pronounced.

\begin{table}[h]
\centering
\caption{Perplexity on WikiText2 comparison for different model with and without BOT quantized (per tensor quantization)}
\begin{tabular}{||l|c|c|c||}
\hline
\textbf{Model} & \textbf{FP16} & \textbf{W8A8 w/ BOT} & \textbf{W8A8 w/o BOT}  \\
\hline
\hline
LLaMA3 8b & 6.14 & 59.9 & 9.02 \\
\hline
LLaMA2 7b & 5.47 & 9.89 & 9.05 \\
\hline
LLaMA2 13b & 4.88 & 57.36 & 8.93 \\
\hline
Mistral 7b & 5.25 & 80.53 & 61.96 \\
\hline
\end{tabular}

\label{tab:BOT_exp}
\end{table}

\begin{figure}[ht]
    \begin{subfigure}[b]{0.44\textwidth}
         \centering
         \includegraphics[width=\textwidth]{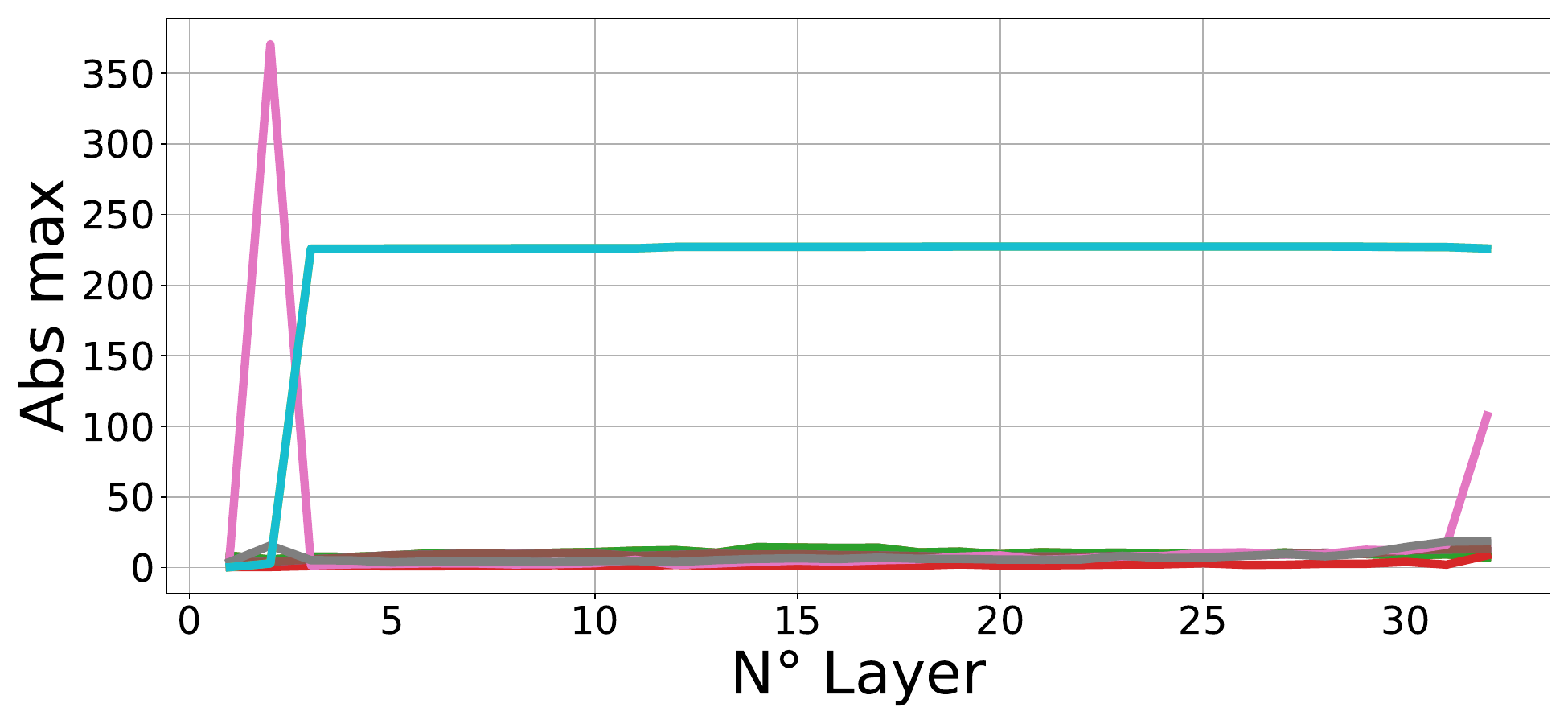}
         \caption{Input}
         \label{fig:LLaMA8B_input_abs}
     \end{subfigure}
     \hfill
     \begin{subfigure}[b]{0.55\textwidth}
         \centering
         \includegraphics[width=\textwidth]{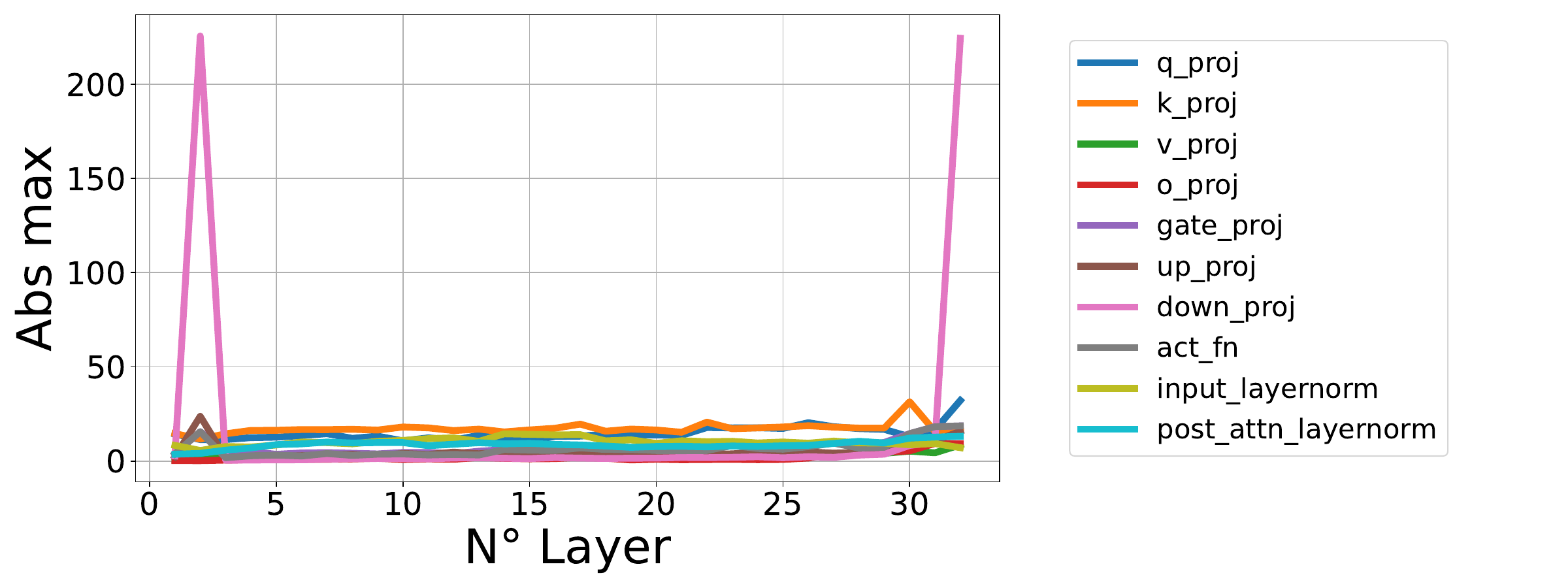}
         \caption{Output}
         \label{fig:LLaMA8B_output_abs}
     \end{subfigure}
     \caption{Maximum absolute value over layers for a LLaMA3-8B. Each color represent a different projection and we clearly see that down\_proj has the biggest spikes in input and output. We also observe that RMSNorm propagate spikes through the entire model.}
        \label{fig:Activations_abs}
\end{figure}

\subsection{Localisation of outliers}

Another crucial question concerns the location of these spikes and their prevalence throughout the model. The answer varies between older and newer models. In OPT-like architectures, as noted in \cite{dettmers2022gpt3}, spikes are absent in the second projection of the FFN. However, this is not the case for LLaMA, as observed in \cite{ahmadian2023intriguing}.

Figure \ref{fig:Activations_abs} corroborates this phenomenon, clearly showing that the down projection exhibits the highest spikes in both the second and final layers of the model. Furthermore, Figure \ref{fig:LLaMA8B_input_abs} demonstrates that spikes are present across all layers of the input RMSNorm and post-attention RMSNorm.

This observation indicates that the residual layer immediately preceding RMSNorm (Figure \ref{fig:Architectures}) propagates spikes throughout the model. This pattern is consistent across LLaMA2, and Mistral7B models (see Appendix \ref{appendix:A}).

These findings highlight the architectural differences between model families and their impact on spike distribution, emphasizing the importance of model-specific considerations in addressing quantization challenges.

\subsection{Architecture and training dependency}

Contemporary LLMs are predominantly based on causal architectures such as GPT, OPT, or LLaMA (Figure \ref{fig:Architectures}), as opposed to encoding-decoding models like BERT. While different architectures lead to varying outlier patterns, some consistencies persist, such as the presence of spikes in layer normalization inputs and the propagation of spikes through residual layers.

A notable difference we've observed is the impact of normalization on spike amplitude. Contrary to the common understanding that normalization layers generate outliers and spikes \cite{xiao2023smoothquant}, we find the opposite effect in LLaMA architectures (Figures \ref{fig:LLaMA8B_input_abs}, \ref{fig:LLaMA8B_output_abs}). The RMS Norm output actually reduces the amplitude of input spikes, which were initially quite high. This pattern is consistent across all LLaMA architectures, as well as Mistral models (Appendix \ref{appendix:A}).

Training is also known to influence the emergence of spikes at inference. As observed in \cite{ahmadian2023intriguing} hyper-parameters like weight decay, gradient clipping, encoding format or drop out can reduce the presence of outliers in LLMs.

To summarize the behavior in LLaMA architectures: spikes are initially generated by the down projection of the second layer, then propagated through the model via residual layers. Subsequently, normalization reduces their amplitude until the final down projection. Interestingly, this pattern doesn't hold for Mistral architectures, despite being based on LLaMA. This difference could be explained by the way Mistral is trained which is known to cause more or less outliers at inference \cite{ahmadian2023intriguing}.

\section{Mixed Precision LLaMA }

Our study focuses on LLaMA-like models, which exhibit numerous invariant features as previously discussed. We will primarily leverage the observation that spikes occur predominantly in two projection layers to develop a mixed-precision quantization pipeline.

Furthermore, we intend to use a different coding format for these layers exhibiting spikes like FP8 which allow a wide range of values and a pretty good precision. This way we can quantize all our linear layers in 8 bits without a significant drop of performance

This targeted strategy aims to optimize quantization for LLaMA-like architectures by addressing the specific characteristics of their spike distribution and leveraging the resilience of these models to quantization.

\subsection{Mixed Precision}

As previously illustrated in Figure \ref{fig:Activations_abs} and detailed in Appendix \ref{appendix:A}, the LLaMA architecture exhibits activation spikes in the down projections (see Figure \ref{fig:LLaMA}) of only two layers: one at the beginning and one at the end of the model. Our proposed approach addresses this by treating these two projections distinctly. We compute these specific projections in FP16 precision, while the rest of the model's operations are executed in lower precision. This strategy enables efficient quantization of activations without resorting to specialized outlier-specific quantizers like SmoothQuant or LLM.int8(), or computing certain tokens in FP16 as in CushionCache \cite{son2024prefixing}.

Our mixed precision pipeline is designed to show that standard absolute maximum uniform quantization can be applied to nearly all layers, with only 2 or 3 projections requiring special treatment. Specifically, Mistral 7B, LLaMA2-7B and LLaMA2-13B models display spikes in the final output projection of their attention mechanism (as shown in Figures \ref{fig:LLaMA2-13B_output_abs} and \ref{fig:Mistral-7B_output_abs}). For these architectural designs, we decided to keep this specific projection in full precision when spike values exceed 100, exempting it from quantization. Table \ref{tab:SumUpQuant} provides an overview of all the layers we maintained in full precision within our mixed-precision approach.

\begin{table}[h]
\centering
\caption{Summarize of layers not quantized for down and out projections}
\begin{tabular}{||c|c|c|c|c|c||}
\hline
\textbf{Projection} & LLaMA3-8B & LLaMA2-7B & LLaMA2-13B & Mistral-7B\\
\hline
\hline
Down & 2, 32 & 2, 31 & 4, 39 & 2, 31, 32  \\
\hline
Out & - & 32 & 40 & 32  \\
\hline
\end{tabular}

\label{tab:SumUpQuant}
\end{table}

\subsection{FP8 format}

To further enhance our quantization pipeline, we used the FP8 format \cite{micikevicius2022fp8} for down projections exhibiting spikes. PyTorch offers two variants of FP8: E5M2 and E4M3, denoting the number of bits allocated to the exponent and mantissa respectively. This format is particularly well-suited for our needs as it enables non-uniform quantization, providing high precision for small values while still accommodating large values. In our specific case, we need to manage significant spiking values that can reach up to 2500 (as shown in Figure \ref{fig:Activations_abs_LLaMA2-7B}). Consequently, we opted for the E5M2 format, which has a maximum representable value of 57344, rather than E4M3, which is limited to a maximum of 448.

\begin{table}[h]
\centering
\caption{Perplexity on WikiText2 for different model with a naive quantization, SmoothQuant and our MixPrecision LLM}
\begin{tabular}{||c|c|c|c|c|c|c|c||}
\hline
\textbf{Bits} & \textbf{Type} & \textbf{Smooth/Mix} & LLaMA3-8B & LLaMA2-7B & LLaMA2-13B & Mistral-7B\\
\hline
\hline

16 & - & - & 6.136 & 5.47 & 4.88 & 5.25\\
\cline{1-7}
\multirow{6}{*}{8} & \multirow{3}{*}{\centering per\_tensor} & - & 40.454 & 8.82 & 50.22 & 80.58\\
& & Smooth & 44.735 & 9.90 & 57.36 & 80.53\\
& & Mix & \textbf{8.24} & \textbf{6.27} & \textbf{8.38} & \textbf{10.14}\\
\cline{2-7}
& \multirow{3}{*}{\centering per\_token} & - & 6.270 & 5.58 & 4.95 & 5.32\\
& & Smooth & 6.266 & 5.59 & 4.94 & \textbf{5.30}\\
& & Mix & \textbf{6.250} & \textbf{5.57} & \textbf{4.94} & 5.32\\
\cline{1-7}
\multirow{6}{*}{6} & \multirow{3}{*}{\centering per\_tensor} & - & 666.033 & 525.18 & 4341.79 & 431.06\\
& & Smooth & 1343.003 & 393.42 & 5045.24 & 1617.82\\
& & Mix & \textbf{196.751} & \textbf{79.4} & \textbf{2046} & \textbf{241.27}\\
\cline{2-7}
& \multirow{3}{*}{\centering per\_token} & - & 7.857 & 6.94 & 5.71 & 6.70\\
& & Smooth & 7.825 & 6.92 & \textbf{5.37} & 8.63\\
& & Mix & \textbf{7.461} & \textbf{6.84} & 5.67 & \textbf{6.1}\\
\hline
\end{tabular}

\label{tab:PPL}
\end{table}

\begin{table}[h]
\centering
\caption{Average Zero Shot accuracy on 7 benchmark for different model with a naive quantization, SmoothQuant and our MixPrecision LLM}
\begin{tabular}{||c|c|c|c|c|c|c|c||}
\hline
\textbf{Bits} & \textbf{Type} & \textbf{Smooth/Mix} & LLaMA3-8B & LLaMA2-7B & LLaMA2-13B & Mistral-7B\\
\hline
\hline

16 & - & - & 67.89 & 64.85 & 67.60 & 68.23\\
\cline{1-7}
\multirow{6}{*}{8} & \multirow{3}{*}{\centering per\_tensor} & - & 51.52 & 57.36 & 51.02 & 46.06\\
& & Smooth & 51.54 & 57.15 & 50.73 & 45.62\\
& & Mix & \textbf{65.46} & \textbf{62.87} & \textbf{60.26} & \textbf{59.36}\\
\cline{2-7}
& \multirow{3}{*}{\centering per\_token} & - & \textbf{67.69} & 63.83 & 67.05 & \textbf{67.74}\\
& & Smooth & 67.34 & 63.96 & \textbf{67.33} & 67.72\\
& & Mix & 67.24 & \textbf{64.52} & 66.62 & 67.66\\
\cline{1-7}
\multirow{6}{*}{6} & \multirow{3}{*}{\centering per\_tensor} & - & 37.11 & 37.5 & 36.93 & 40.49\\
& & Smooth & 38.38 & 38.9 & 36.76 & 36.85\\
& & Mix & \textbf{39.23} & \textbf{45.68} & \textbf{37.04} & \textbf{44.5}\\
\cline{2-7}
& \multirow{3}{*}{\centering per\_token} & - & 62.84 & 61.39 & 62.93 & 64.15\\
& & Smooth & \textbf{63.26} & \textbf{63.46} & \textbf{64.37} & \textbf{66.24}\\
& & Mix & 63.11 & 61.5 & 63.79 & 64.15\\
\hline
\end{tabular}

\label{tab:AZS}
\end{table}

\section{Experiments}

\subsection{Setups}

Our experiments were made on 4 different architectures : LLaMA2-7B, LLaMA2-13B, LLaMA3-8B and Mistral7B and we compared our results with a very used state of the art quantization scheme : SmoothQuant and a naive uniform quantization.

To compare our performances we computed the perplexity of the predictions on WikiText2 and the average accuracy on 7 zero shot benchmarks : Winogrande, Lambada, RTE, OpenBookQA, PIQA, HellaSwag and COPA. To perform the evaluation of our model we used LM evaluation harness benchmark by EleutherAI \cite{lintang_sutawika_2023_10256836}. We used a context size of 2048 tokens for all our experiments.

Our quantization strategy is selectively applied to specific components of the model. We implement quantization in linear layers, including the Query-Key-Value projections and the Multi-Layer Perceptron (MLP) components. However, we maintain full precision for certain critical operations: the matrix multiplication and softmax function within the attention mechanism, as well as the Root Mean Square (RMS) normalization. This targeted approach allows us to balance efficiency gains from quantization with preservation of model accuracy in key computational steps.

\subsection{Results}

\subsubsection{Unquantized Down and Out Projections}.
Tables \ref{tab:PPL}  presents the perplexity (PPL) results for our mixed precision method (excluding FP8). Our approach consistently outperforms perplexity of alternative schemes, particularly in 8-bit per-tensor quantization, where we observe a PPL reduction exceeding 30 for LLaMA3-8B. While 6-bit quantization remains somewhat unstable, our method still achieves applicative performances despite elevated perplexity. Per-token quantization proves less challenging for all methods, yet our approach maintains its edge in this scenario as well.

About the average zero shot accuracy Table \ref{tab:AZS} our mixed precision method is very efficient in 8 and 6 bits per tensor where we can increase accuracy over 10 points. Per token quantization is slightly different, in 8 bits all methods almost reach FP16 baseline and in 6 bits it appears to be dominated by SmoothQuant method and not ours even if PPL is lower.

These findings support our primary hypothesis that in LLaMA-like models, spikes are predominantly concentrated in two or three specific locations. This allows us to quantize the remaining portions effectively without considering outliers, while maintaining performance integrity.

\subsubsection{Random Mixed Precision}. To further validate our hypothesis that the outlier problem is primarily associated with specific down and out projections, we conducted an experiment where we randomly apply FP16 to different down and out projections from Table \ref{tab:SumUpQuant}. Table \ref{tab:RandMix} presents a comparison of perplexity between random and targeted mixed precision approaches, averaged across three different seeds.

The results clearly demonstrate that applying higher precision to random projections, rather than targeting those known to contain spikes, does not yield performance improvements. This finding strongly supports our conclusion that the quantization challenge in LLaMA models is primarily confined to three specific linear projections, rather than being a model-wide issue.
\begin{table}[h]
\centering
\caption{Comparison of Perplexity between targeted Mixed Precision with Random layer Mixed Precision in 8 bits per tensor averaged on 3 different seeds}
\begin{tabular}{||c|c|c|c|c|c|c|c||}
\hline
\textbf{Bits} & \textbf{Type} & \textbf{Mix/Rand} & LLaMA3-8B & LLaMA2-7B & LLaMA2-13B & Mistral-7B\\
\hline
\hline
\cline{1-7}
\multirow{2}{*}{8} & \multirow{2}{*}{\centering per\_tensor}
& Mix & 8.24 & 6.28 & 8.38 & 10.14 \\
& & Mix + Rand & $39.76 \pm 0.24$ & $8.68 \pm 0.12$ & $49.28 \pm 0.46$ & $78.82 \pm 1.14$\\
\cline{1-7}

\hline
\end{tabular}

\label{tab:RandMix}
\end{table}

\subsubsection{FP8 Down Projections}

We now test FP8 format on down and out projections that were previously excluded from quantization. Table \ref{tab:FP8} shows the results on WikiText2 perplexity in 8 bits quantization and we can see that most of the time using FP8 format doesn't injure performances and can even outperform previous capacities like for LLaMA3-8B per token which is better by 1 point of PPL than using FP16 format. Research has demonstrated that full model quantization to FP8 is highly effective for BERT models \cite{shen2024efficient}, and this approach appears to be similarly successful when applied to LLaMA models (FP8 row of Table \ref{tab:FP8}). While FP8 quantization is beneficial, 8-bit integer (INT8) operations offer even faster computation. This suggests that combining both FP8 and INT8 quantization techniques could be particularly advantageous for optimizing model performance and efficiency.

These results show the potential of these techniques on LLaMA-like architectures. By only treating 2 or 3 projections in FP8 we can outperform quantization techniques that were specifically designed to handle spikes and outliers for all types of architectures. We showed that the problem of spikes is relevant only in specific locations and that we don't need a general method to outperform previous techniques.

\begin{table}[h]
\centering
\caption{Comparison of Perplexity between Mixed Precision with and without FP8 format for down and out projections}
\begin{tabular}{||c|c|c|c|c|c|c|c||}
\hline
\textbf{Bits} & \textbf{Type} & \textbf{Mix/FP8} & LLaMA3-8B & LLaMA2-7B & LLaMA2-13B & Mistral-7B\\
\hline
\hline
\multicolumn{3}{|c|}{FP8} & 6.42 & 5.58 & 4.98 & 5.35  \\
\hline
\hline
\cline{1-7}
\multirow{4}{*}{8} & \multirow{2}{*}{\centering per\_tensor}
& Mix & 8.24 & 6.28 & 8.31 & 9.97\\
& & Mix + FP8 & 10.64 & 6.44 & 8.50 & 8.23\\
\cline{2-7}
& \multirow{2}{*}{\centering per\_token}
& Mix & 6.250 & 5.57 & 4.94 & 5.32\\
& & Mix + FP8 & 5.32 & 5.57 & 4.95 & 6.26\\
\cline{1-7}

\hline
\end{tabular}
\label{tab:FP8}
\end{table}

\section{Conclusion}

In this paper, we have presented a comprehensive analysis of activation outliers in LLMs, with a specific focus on the LLaMA architecture. Our investigation challenges some existing assumptions about the nature and distribution of outliers in LLMs, revealing that their behavior can vary significantly across different model architectures.

We proposed a novel mixed-precision quantization approach tailored for LLaMA-like models, leveraging the observation that spikes predominantly occur in specific projection layers. By treating these layers separately with higher precision (FP16 or FP8) while quantizing the rest of the model to lower bit-widths, we achieved superior performance compared to existing quantization methods.

Our experimental results demonstrate the effectiveness of this approach across various LLaMA-based models, including LLaMA2, LLaMA3, and Mistral. We observed significant improvements in perplexity and zero-shot accuracy, particularly for 8-bit per-tensor quantization. The method also showed promise for more aggressive 6-bit quantization, though with some instability.

The success of our mixed-precision approach underscores the importance of architecture-specific quantization strategies. By focusing on the unique characteristics of LLaMA-like models, we were able to develop a more efficient quantization pipeline that outperforms general-purpose methods designed to handle outliers across all types of architectures.

Future work could explore the application of this method to other model families, investigate the causes of the remaining instability in 6-bit quantization, and further optimize the use of FP8 format for critical layers. Additionally, combining our approach with other quantization techniques, such as SmoothQuant, could potentially yield even greater performance improvements.

\begin{credits}
\subsubsection{\discintname}
The authors have no competing interests to declare that are relevant to the content of this article
\end{credits}

\newpage
\section*{Appendix}
\appendix
\section{Spikes in LLaMA like architectures}\label{appendix:A}

\begin{figure}[ht]
    \begin{subfigure}[b]{0.44\textwidth}
         \centering
         \includegraphics[width=0.95\textwidth]{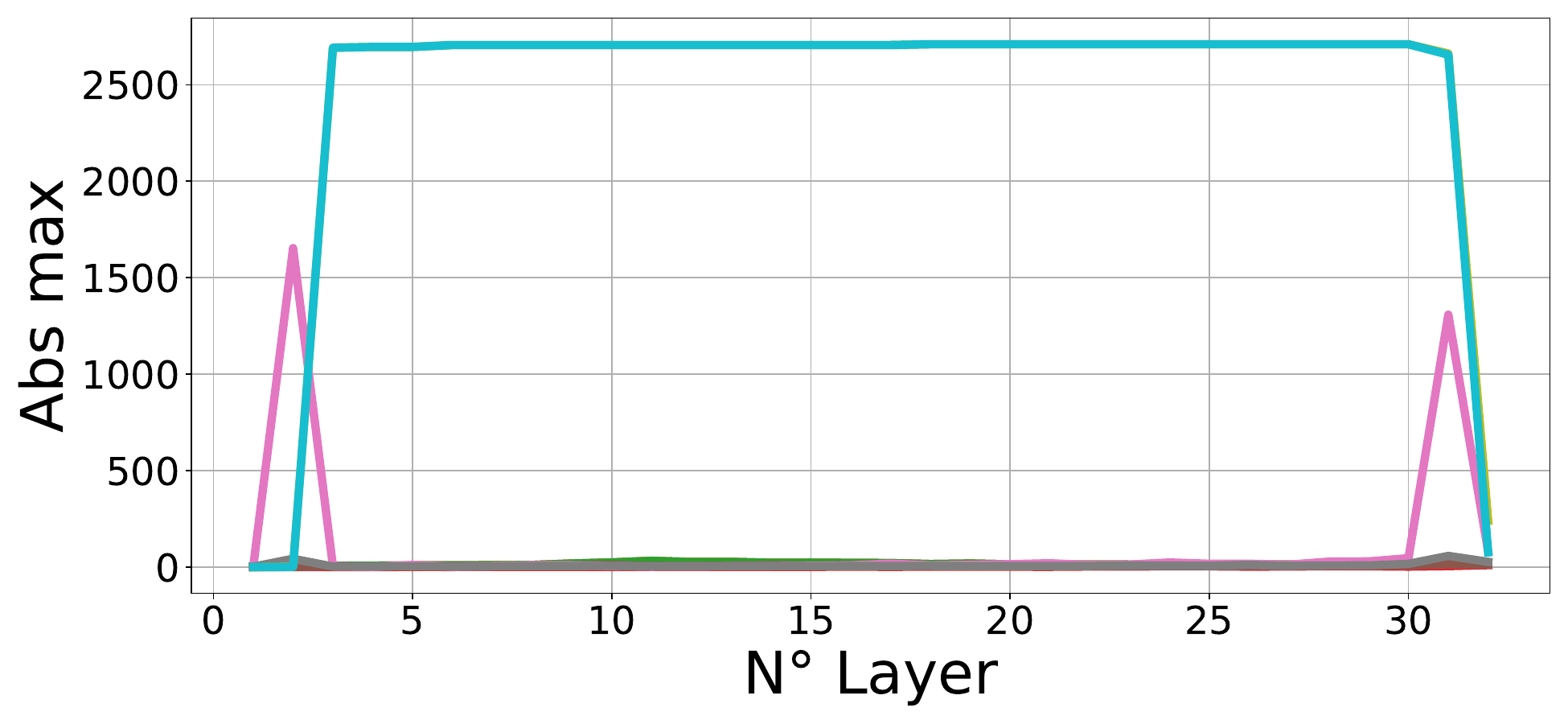}
         \caption{Input}
         \label{fig:LLaMA2-7B_input_abs}
     \end{subfigure}
     \hfill
     \begin{subfigure}[b]{0.55\textwidth}
         \centering
         \includegraphics[width=0.95\textwidth]{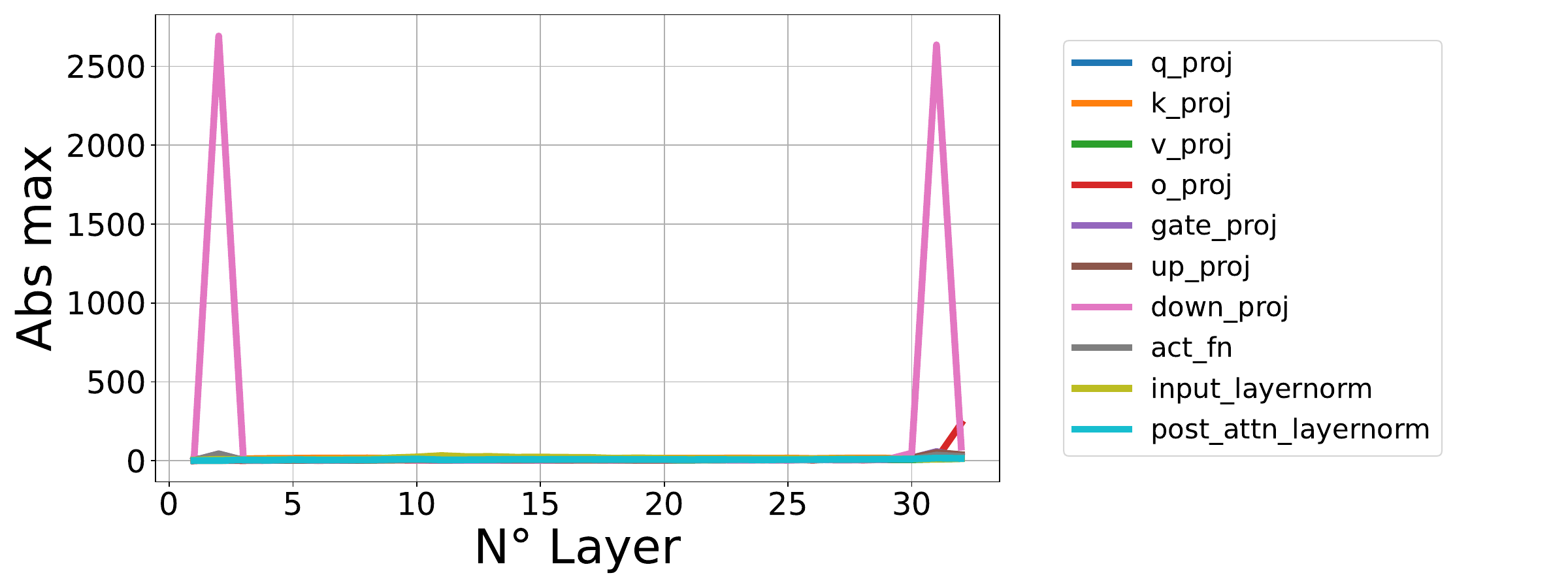}
         \caption{Output}
         \label{fig:LLaMA2-7B_output_abs}
     \end{subfigure}
     \caption{Maximum absolute value over layers for a LLaMA2-7B.}
        \label{fig:Activations_abs_LLaMA2-7B}
\end{figure}

\begin{figure}[ht]
    \begin{subfigure}[b]{0.44\textwidth}
         \centering
         \includegraphics[width=0.95\textwidth]{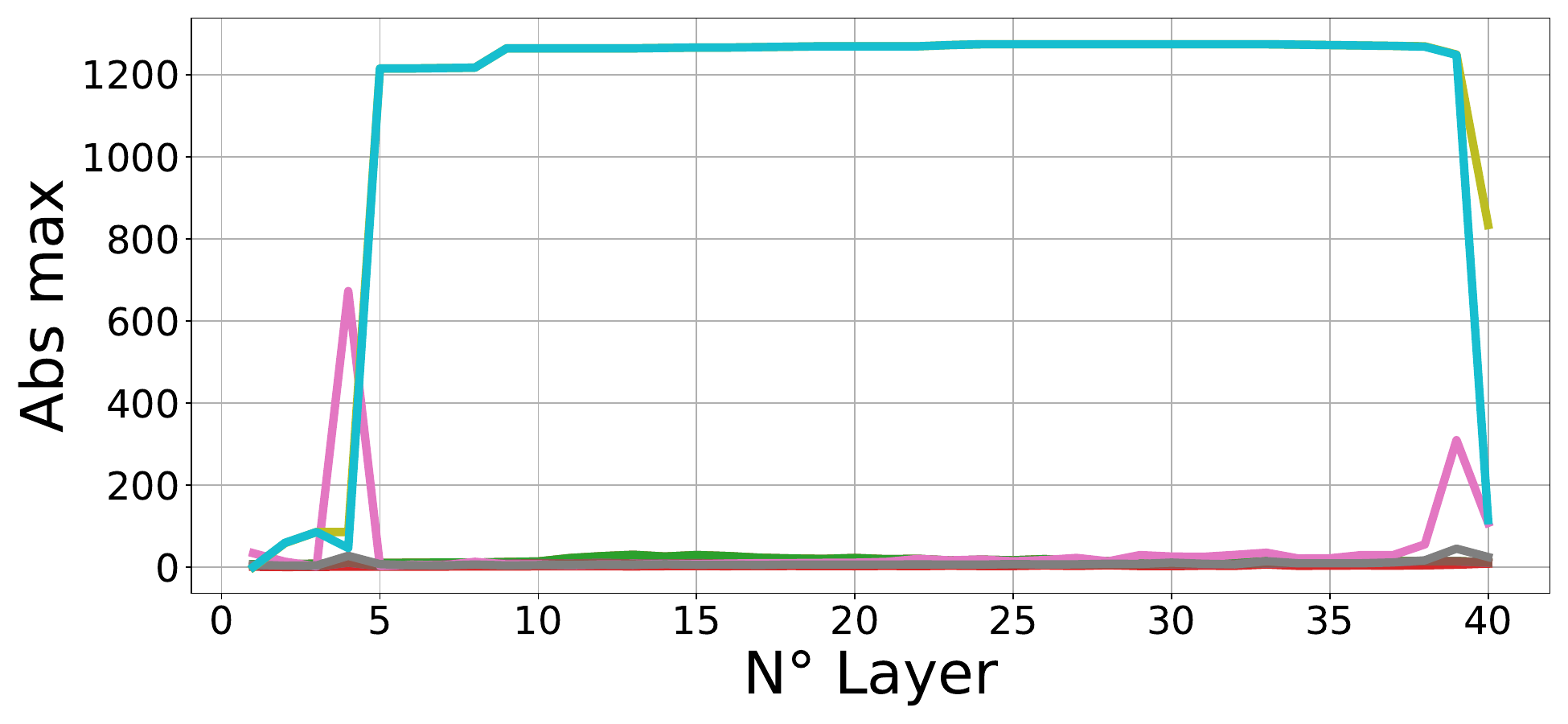}
         \caption{Input}
         \label{fig:LLaMA2-13B_input_abs}
     \end{subfigure}
     \hfill
     \begin{subfigure}[b]{0.55\textwidth}
         \centering
         \includegraphics[width=0.95\textwidth]{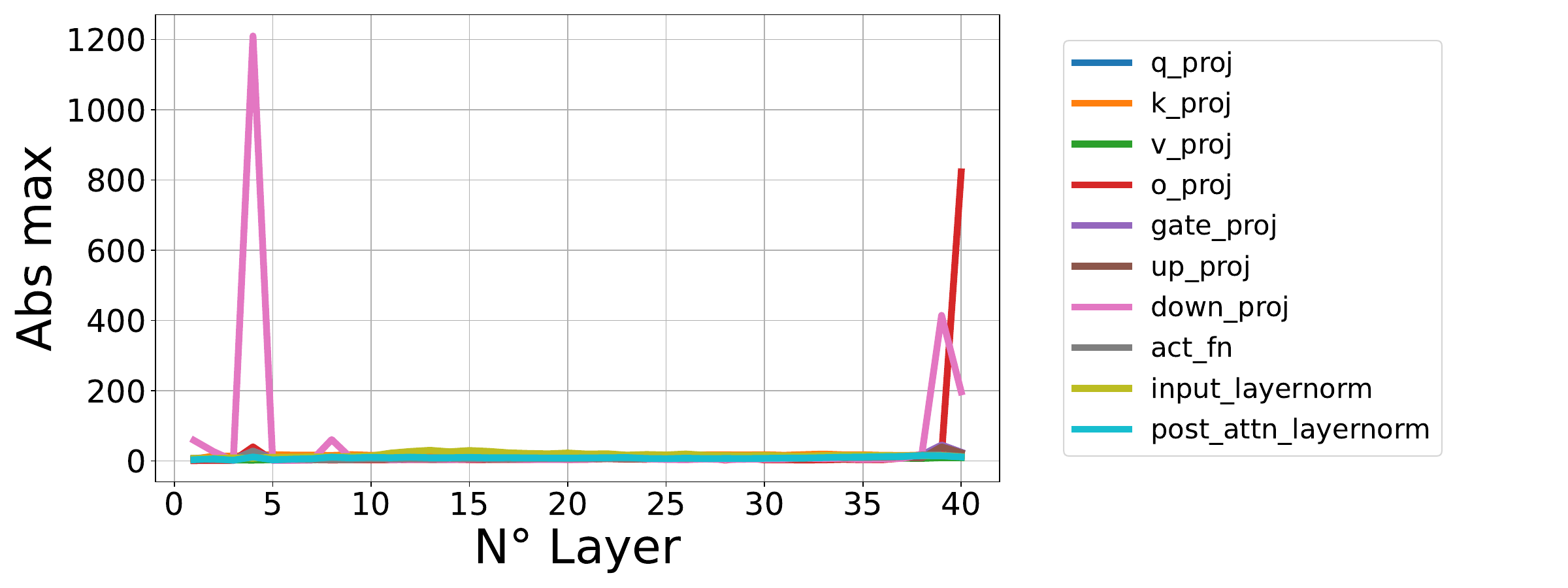}
         \caption{Output}
         \label{fig:LLaMA2-13B_output_abs}
     \end{subfigure}
     \caption{Maximum absolute value over layers for a LLaMA2-13B.}
        \label{fig:Activations_abs_LLaMA2-13B}
\end{figure}

\begin{figure}[ht]
    \begin{subfigure}[b]{0.44\textwidth}
         \centering
         \includegraphics[width=0.95\textwidth]{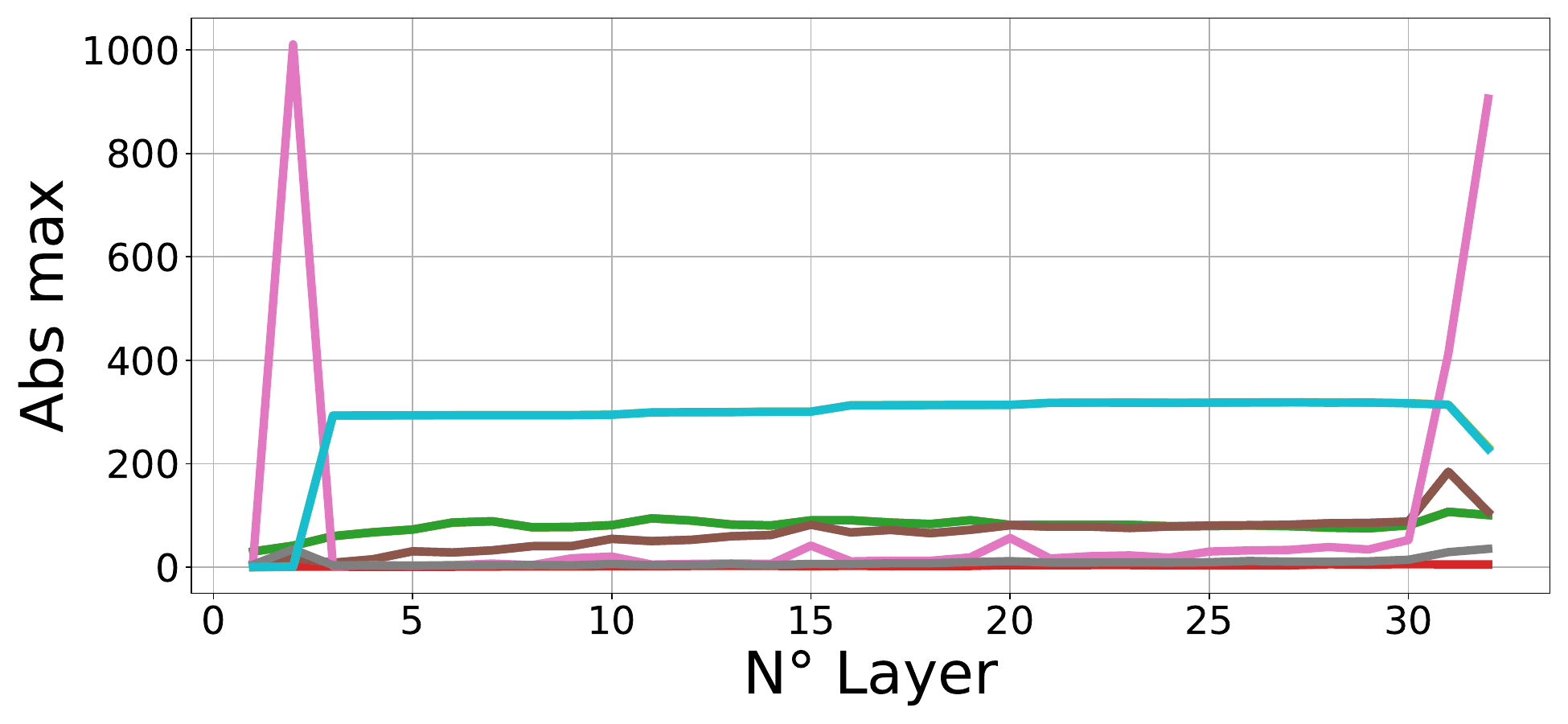}
         \caption{Input}
         \label{fig:Mistral-7B_input_abs}
     \end{subfigure}
     \hfill
     \begin{subfigure}[b]{0.55\textwidth}
         \centering
         \includegraphics[width=0.95\textwidth]{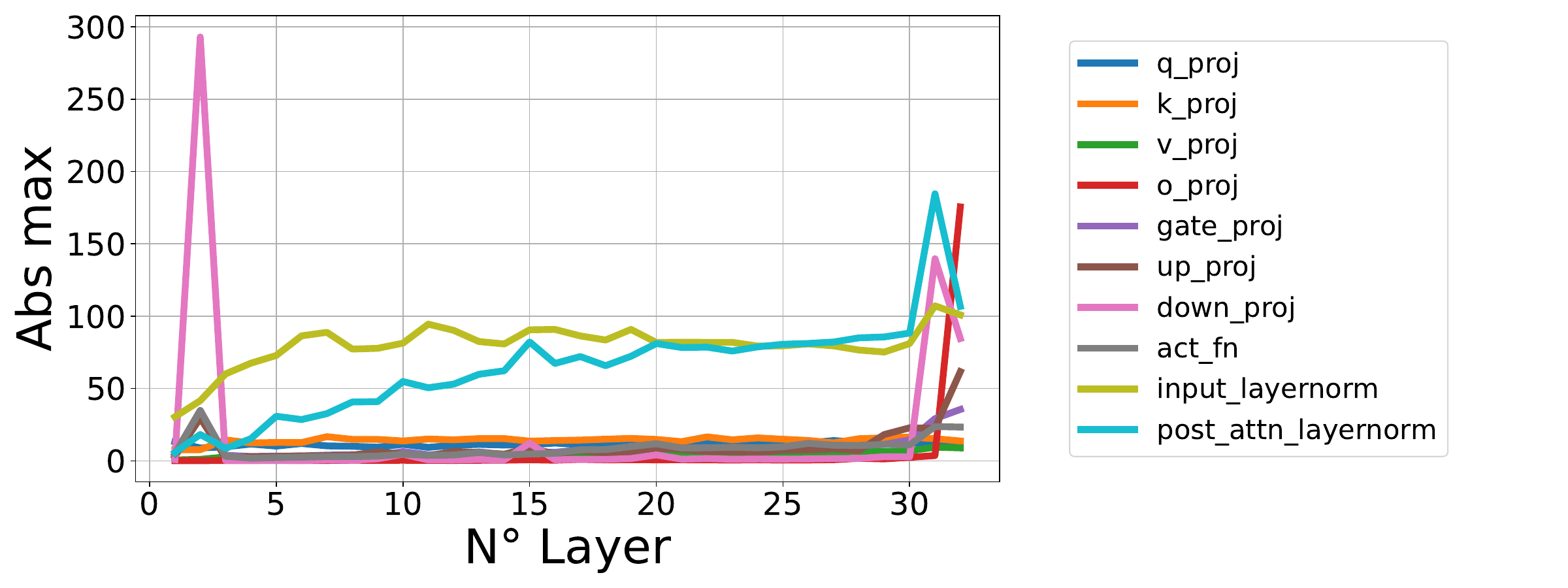}
         \caption{Output}
         \label{fig:Mistral-7B_output_abs}
     \end{subfigure}
     \caption{Maximum absolute value over layers for a Mistral-7B.}
        \label{fig:Activations_abs_Mistral-7B}
\end{figure}

\bibliographystyle{splncs04}
\bibliography{paper}

\end{document}